\documentclass{article} 
\usepackage{nips15submit_e,times}
\usepackage{hyperref}
\usepackage{url}
\usepackage{makecell}
\usepackage{tablefootnote}
\usepackage{algorithm}
\usepackage{algorithmic}
\usepackage{amsmath}
\usepackage{amsfonts}
\usepackage{mathrsfs}
\usepackage{amsthm}

\usepackage[dvips]{graphicx} 

\title{Fast Riemannian-manifold Hamiltonian Monte Carlo for hierarchical Gaussian-process models}

\author{
Takashi Hayakawa \  and Satoshi Asai \\
School of Medicine\\
Nihon University\\
Tokyo, Japan, 1738610 \\
\texttt{hayakawa.takashi@nihon-u.ac.jp} \\
}

\nipsfinalcopy 
\begin{document}

\maketitle

\begin{abstract}
Hierarchical Bayesian models based on Gaussian processes are considered useful for describing complex nonlinear statistical dependencies among variables in real-world data. However, effective Monte Carlo algorithms for inference with these models have not yet been established, except for several simple cases. In this study, we show that, compared with the slow inference achieved with existing program libraries, the performance of Riemannian-manifold Hamiltonian Monte Carlo (RMHMC) can be drastically improved by optimising the computation order according to the model structure and dynamically programming the eigendecomposition. This improvement cannot be achieved when using an existing library based on a naive automatic differentiator. We numerically demonstrate that RMHMC effectively samples from the posterior, allowing the calculation of model evidence, in a Bayesian logistic regression on simulated data and in the estimation of propensity functions for the American national medical expenditure data using several Bayesian multiple-kernel models. These results lay a foundation for implementing effective Monte Carlo algorithms for analysing real-world data with Gaussian processes, and highlight the need to develop a customisable library set that allows users to incorporate dynamically programmed objects and finely optimises the mode of automatic differentiation depending on the model structure.  
\end{abstract}







\section{Introduction}
Bayesian analysis based on Gaussian processes (GPs) associated with positive-semidefinite kernels has been studied extensively from modelling, algorithmic, and theoretical points of view \cite{Williams2006, vanderVaart2008, vanderVaart2011, Suzuki2012, Frigola2013, Svensson2016, Yerramilli2023, Finocchio2023}. A GP serves as a prior process for generating nonlinear functions that describe relationships between variables. The goal of Bayesian analysis is to infer the posterior process for such nonlinear functions conditioned on given data or to estimate the average of quantities of interest over such a posterior process. Although the posterior process can be represented as a closed-form solution for the simplest problems (e.g., Chapter 6 of Ref. \cite{Bishop2006}), such a solution cannot be expected when GPs are used as building blocks for hierarchical models that capture complex statistical dependencies among variables in real-world data. Algorithms for inference with hierarchical GP models can be broadly divided into variational algorithms and Monte Carlo (MC) algorithms. Variational algorithms (e.g., Refs.\cite{Frigola2013, Yerramilli2023}) are usually much faster than MC algorithms but introduce unavoidable bias due to discrepancies between the original and variational models. In contrast, MC algorithms (e.g., Refs.\cite{Svensson2016, Paquet2018, Pandita2021}) allow for precise computation of averages over the posterior, provided the algorithm is run until convergence. The preferred algorithm type depends on the purpose of the computation.

With biometric and econometric applications in mind, we seek an efficient MC algorithm in this study. In these applications, biased results can be misleading and may adversely affect public health or the economy; therefore, the use of variational algorithms is not appropriate. However, to our knowledge, a standard MC algorithm for general hierarchical GP models has not yet been established. Svensson et al. \cite{Svensson2016} sampled a function from the posterior process using a closed-form solution for the GP component of their model conditioned on other parameters. Their formula can be applied only when the likelihood term is Gaussian and is thus not applicable to, for example, binary classification. Hensman et al. \cite{Hensman2015} introduced a non-Gaussian variational approximation of the posterior with the aid of inducing points sampled using Hamiltonian MC (HMC). Their method, however, is variational and not guaranteed to describe the posterior precisely, and its scalability with increasing numbers or dimensions of GPs remains unclear. Pandita et al. \cite{Pandita2021} introduced an adaptive sequential MC for GPs to solve problems in mechanical engineering. However, the authors used hundreds of CPU cores for inference, and the efficiency of their sampling scheme itself was not clearly demonstrated.

One fundamental issue in MC sampling from the posterior in a GP model is that the posterior may be highly stretched or compressed along unknown directions in the parameter space used for sampling. To address this problem, Paquet and Fraccaro \cite{Paquet2018} implemented Riemannian-manifold HMC (RMHMC) for GP models. With the Hessian metric they employed, RMHMC adjusts the direction of MC moves according to the curvature of the log-posterior density surface. They demonstrated that RMHMC is much more efficient than the ordinary HMC method based on the Euclidean metric. However, their method is applicable only to cases with log-concave posterior densities, whereas hierarchical models that describe complex dependencies among variables typically have non-log-concave posterior densities. To overcome this limitation, we develop an efficient implementation of RMHMC for hierarchical GP models using a soft-absolute Hessian metric introduced by Betancourt \cite{Betancourt2013a}. This metric transforms an indefinite Hessian into a positive-semidefinite matrix that shares eigenvectors with the Hessian but has associated eigenvalues set close to the absolute values of the original Hessian eigenvalues. Although a general-purpose implementation of RMHMC with this metric \cite{Cobb2019} based on PyTorch \cite{PyTorch} is available, its performance is poor. We identify the reasons for this, and show that the computational complexity of the implementation can be considerably reduced.

The article is organised as follows. In section \ref{sec2.1}, we introduce our problem setting in which multiple GPs are used as building blocks for a hierarchical model, employing the representation of GPs introduced by Solin et al. \cite{Solin2020}. In section \ref{sec2.2}, we provide a brief introduction to RMHMC with a soft-absolute Hessian metric. In sections \ref{sec2.3} and \ref{sec2.4}, we identify two sources of redundancy in the calculation of the gradient flow and metric, respectively. We show how performance deteriorates when reverse-mode automatic differentiation and a divide-and-conquer algorithm for the eigendecomposition of symmetric matrices in PyTorch are naively applied. In section \ref{sec3}, we numerically demonstrate that our implementation considerably outperforms an implementation based on a general-purpose library. Specifically, in section \ref{sec3.1}, we compare the performance of different implementations using toy examples of varying sizes and show that both sources of redundancy contribute to the improvement. In section \ref{sec3.2}, we compare the performance of our implementation with that of no-U-turn (NUT)-HMC based on the Euclidean metric. We show that NUT-HMC becomes trapped within a very narrow region and fails to converge to the posterior, suggesting that the use of a data-adaptive metric is essential for efficient sampling. In section \ref{sec3.3}, we show, using a toy example and a real-world dataset, that our implementation successfully calculates the marginal likelihood, (that is, the Bayesian model evidence) within a reasonable computation time. Finally, in section \ref{sec4}, we discuss, based on these results, the advantages and limitations of the proposed method compared with existing approaches and suggest directions for future work. Mathematical notations used in this article are summarised in Table \ref{table2}.

\section{Methods and Algorithms}
\subsection{Hierarchical GP models and their approximate representations} \label{sec2.1}
In this study, we consider hierarchical GP models for which the samplewise likelihood of {\it i.i.d.} data $X_i \in \mathbf{R} ^d$ (with sample index $i\in \{ 1,2,\ldots ,N\} $) is parameterised by $J$ nonlinear functions $\{ f_j(X_i)\} _{1\leq j\leq J}$ as
\begin{eqnarray}
P(X_i|\{ f_j\} _{1\leq j\leq J})=\exp(-U(\{ f_j(X_i)\} _{1\leq j\leq J})). \label{data_likelihood}
\end{eqnarray}
For instance, the simplest examples describing the relationship between a target variable $X_i^{(\mathrm{tgt})}$ and covariates $X_i^{(\mathrm{cov})}$ are as follows.

{\bf Example 1. Covariate-dependent target mean and variance ($J=2$):}  
\begin{eqnarray}
X_i^{(\mathrm{tgt})}=f_1(X_i^{(\mathrm{cov})})+\epsilon _i, \ \  \epsilon _i \sim \mathcal{N} (0, \delta +\exp f_2(X_i^{(\mathrm{cov})})), \label{mvmodel}
\end{eqnarray}
with fixed $\delta (>0)$.

{\bf Example 2. Nonlinear Gaussian mixture model ($J=4$):}
\begin{eqnarray}
X_i^{(\mathrm{tgt})}=(1-s_i)f_1(X_i^{(\mathrm{cov})})+s_if_2(X_i^{(\mathrm{cov})})+\epsilon _i, \nonumber \\
\epsilon _i\sim \mathcal{N} (0, \delta +\exp f_3(X_i^{(\mathrm{cov})})) ,\ \ \ s_i\sim \mathrm{Bernoulli} [1/(1+\exp f_4(X_i^{(\mathrm{cov})}))].
\end{eqnarray}

In the above modelling, we assume that each of $\{ f_j\} _{1\leq j\leq J}$ is the sum of nonlinear functions generated by GP priors $\{ \mathcal{G} _{jk}\} _{1\leq j\leq J, k\in \mathcal{K} _j}$:
\begin{eqnarray}
f_j=\sum _{k\in \mathcal{K} _j}f_{jk}+b_j, \  \  f_{jk}\sim \mathcal{G} _{jk}(\{ \theta _{jk\ell} \} _{\ell \in \mathcal{L}_{jk}}), \ \ b_j\sim \mathcal{N} (0, \Sigma ). \label{BMKL}
\end{eqnarray}
Each GP $\mathcal{G} _{jk}$ is associated with a positive-semidefinite kernel $k(\cdot, \cdot )$ that describes the prior covariance of the generated functions and is parameterised by scalar hyperparameters $\{ \theta _{jk\ell }\} _{\ell \in \mathcal{L}_{jk}}\in \mathbf{R} ^{|\mathcal{L}_{jk}|}$. Thus, each of $\{ f_j\} _{j\in \mathcal{J}}$ is a Bayesian multiple-kernel model for which statistical properties were investigated in a previous study \cite{Suzuki2012}.  The prior density of the hyperparameters $\Pi _{jk}(\{ \theta _{jk\ell }\} _{\ell \in \mathcal{L}_{jk}})$ is assumed to be given. 

As we jointly sample functions $\{ f_{jk}\} _{j,k}$ and hyperparameters $\{ \theta _{jk\ell }\} _{j,k,\ell}$, the complicated dependence of the prior density of $\{ f_{jk}(X_i)\} _{i}$ on $\{ \theta _{jk\ell }\} _{j,k,\ell }$ hinders efficient sampling. To alleviate this problem, one previous study \cite{Solin2020} introduced a useful approximation scheme that took the following form: 

\begin{eqnarray}
f_{jk}(X)\approx \sum _{m=1}^{M_{jk}}a_{jkm}\phi _{jkm}(X), \ \ a_{jkm}\sim \mathcal{N} (0, c_{jk}V_{jkm}(\sigma _{jk})), \label{Solin_representation}
\end{eqnarray}      
where the generated function was represented as the sum of sinusoidal feature functions $\phi _{jkm}$ weighted by the normally distributed coefficients $a_{jkm}$. For the Gaussian kernels that we mainly use in this study, hyperparameters $\{ \theta _{jk\ell }\} _{\ell \in \mathcal{L} _{jk}} =\{ c_{jk}, \sigma _{jk}\} $ tune the amplitude and bandwidth, respectively, of the GP through the above equation. Concrete representations of $\phi _{jkm}$ and $V_{jkm}$ are given in Appendix \ref{appendix_Solin}. In this study, for simplicity, we restrict ourselves to the use of one-dimensional Gaussian kernels and linear kernels, as well as the use of a single set of hyperparameters $\{ c_g, \sigma _g\} $ for all the GPs associated with Gaussian kernels and a single hyperparameter $c_{\ell }$ for all the GPs associated with linear kernels (equivalently Gaussian random variables). Note that $V_{jkm}$ can be set to $1$ for linear kernels. We assign inverse Gamma priors to the hyperparameters as follows:
\begin{eqnarray}
\theta \sim \mathrm{InvGamma} (\alpha _{\theta }, \beta _{\theta }), \ \ (\theta =c_g,\ \sigma _g\ \mathrm{or} \ c_\ell ). \label{hpar_prior}
\end{eqnarray}

In the $L, M\rightarrow \infty $ limit (see Appendix \ref{appendix_Solin} for the definition of $L$), the above approximation tends to be exact (Theorem 4 of Ref.\cite{Solin2020}). The advantage of this approximation is the use of features ($\phi _{jkm}$) that are independent of the bandwidth parameter $\sigma _{jk}$, and the use of a tractably computed prior density for $\{ a_{jkm}\} _{j,k,m}$. However, the prefixed sinusoidal features are less flexible than are data-adaptive features (obtained, e.g., by the incomplete Cholesky decomposition \cite{Bach2003}) and are not suitable for approximating high-dimensional functions. In biometric and econometric applications (e.g., epidaemiology), modelling a target variable as the sum of the effects of several factors is preferred for the sake of interpretability and robustness. Thus, modelling with the sum of several low-dimensional functions in Eq.(\ref{BMKL}) is justifiable.

The above modelling motivates our investigation of efficient MC algorithms. The variance $c_{jk}V_{jkm}$ varies with $\sigma _{jk}$ over a few orders of magnitude. Combined with the structure of the data likelihood, the posterior distribution is often extremely stretched or compressed along unknown directions. Thus, a data-adaptive sampling scheme, such as RMHMC, is required.

\subsection{RMHMC with a soft-absolute Hessian metric} \label{sec2.2}
An HMC algorithm \cite{Duane1987, Neal1993, Neal1996} obtains samples of parameters $q\in \mathbf{R} ^d$ from a target density $P(q)$ by regarding $q$ as the position of particles and simulating its time evolution together with that of the associated momentum vector $p\in \mathbf{R} ^d$ according to the Hamilton equation of motion with a suitably designed Hamiltonian $\mathcal{H}(q,p)$. The time evolution over discretised timesteps is determined by  
\begin{eqnarray}
p\left (t+\frac{\epsilon }{2}\right )&=&p(t)-\frac{\epsilon }{2} \frac{\partial }{\partial q}\mathcal{H}\left (q(t), p\left (t+\frac{\epsilon }{2}\right ) \right ), \nonumber \\
q(t+\epsilon )&=&q(t)+\frac{\epsilon }{2} \left \{ \frac{\partial }{\partial p}\mathcal{H}\left (q(t), p\left (t+\frac{\epsilon }{2}\right )\right )+\frac{\partial }{\partial p} \mathcal{H}\left (q(t+\epsilon), p\left (t+\frac{\epsilon }{2}\right )\right )\right \}, \nonumber  \\
p(t+\epsilon )&=&p\left (t+\frac{\epsilon }{2} \right)-\frac{\epsilon }{2} \frac{\partial }{\partial q} \mathcal{H}\left (q(t+\epsilon ), p\left (t+\frac{\epsilon }{2}\right ) \right ). \label{leapfrog}
\end{eqnarray}
This time evolution for a time period of duration $\epsilon $ is called a leapfrog. The mapping from $(q(t), p(t))$ to $(q(t+\epsilon ), p(t+\epsilon ))$ preserves the volume and approximately preserves the value of $\mathcal{H}$. As we show in Algorithm \ref{Algo}, HMC repeats the combination of $C$ leapfrogs and the sampling of a new value for $p$ from a suitable $q$-dependent Gaussian distribution. At the end of every $C$ leapfrogs, the change in $\mathcal{H}$ due to discretisation is adjusted via the Metropolis-Hastings procedure. Samples obtained from a long-time simulation after a burn-in period approximate the target distribution, and the statistics calculated with the samples converge to the average over the target at the infinite time limit. Girolami and Calderhead \cite{Girolami2011} introduced the use of a nontrivial Riemannian metric $q\mapsto G(q)\in \mathbf{R} ^d\times \mathbf{R} ^d$ for the parameter space and showed that, in this case, the Hamiltonian should be defined as
\begin{eqnarray}
\mathcal{H}(q,p)\overset{\mathrm{def}}{=} -\ln P(q)+\frac{1}{2} \ln \left \{ (2\pi )^d|G(q)|\right \}+\frac{1}{2} p^TG(q)^{-1}p. \label{Hamiltonian}
\end{eqnarray}
For this version of Hamiltonian, their derivatives with respect to $q$ and $p$ read
\begin{eqnarray}
\frac{\partial }{\partial p_i}\mathcal{H}&=&\left (G(q)^{-1}p\right )_i \nonumber \\
\frac{\partial }{\partial q_i}\mathcal{H}&=&-\frac{\partial }{\partial q_i} \ln P(q)+\frac{1}{2} p^TG(q)^{-1}\frac{\partial G(q)}{\partial q_i}G(q)^{-1}p-\frac{1}{2} tr\left (G(q)^{-1}\frac{\partial G(q)}{\partial q_i}\right ). \label{d_Hamiltonian}
\end{eqnarray}
The choice of $G(q)=1$ recovers the ordinary HMC based on the Euclidean metric.

\begin{algorithm}[tb]
   \caption{(RM)HMC}
   \label{Algo}
\begin{algorithmic}
   \STATE {\bfseries Input:} Target density $P(q)$, initial value $q(0)\in \mathbf{R} ^d$, number of total and burn-in MC moves $A$ and $A_0$, stepsize $\epsilon $, number of leapfrogs $C$ in a single move, metric $G: q\mapsto G(q)\in \mathbf{R} ^d\times \mathbf{R} ^d$ and Hamiltonian $\mathcal{H} : (q,p)\mapsto \mathcal{H}(q,p)\in \mathbf{R}$ defined in Eq.(\ref{Hamiltonian}).  
   \STATE {\bfseries Output:} A collection of parameter values $\{ q(\epsilon \ell C)\} _{A_0\leq \ell \leq A}$ useful for approximating $P(q)$.
   \STATE Generate an initial value for the momentum vector: $p(0)\sim \mathcal{N}(0, G(q(0)))$.
   \FOR{$\ell $ in $\{0 \ldots A-1\}$}
   \FOR{$s$ in $\{ 0\ldots C-1\} $}
   \STATE Carry out a leapfrog step according to Eq.(\ref{leapfrog})
   \ENDFOR
   \STATE Generate $r_{\ell }\sim \mathrm{Uniform}(0, 1)$ for the following Metropolis-Hastings procedure: 
   \IF{$r_\ell <\exp (-\mathcal{H}(q(\epsilon (\ell+1)C), p(\epsilon (\ell+1)C))+\mathcal{H}(q(\epsilon \ell C), p(\epsilon \ell C)))$}
   \STATE $q(\epsilon (\ell +1)C)\leftarrow q(\epsilon \ell C)$
   \ENDIF
   \STATE Generate an updated value for the momentum vector: $p(\epsilon (\ell +1)C)\sim \mathcal{N}(0, G(q(\epsilon (\ell +1)C)))$.
   \ENDFOR
\end{algorithmic}
\end{algorithm}

The Hessian of $-\ln P(q)$ is a natural choice for $G(q)$, as we explained in the previous section. However, the Hessian of the negative log-posterior density for a hierarchical probabilistic model is often not positive definite and therefore cannot be used as a metric. For this problem, Betancourt \cite{Betancourt2013a} proposed transforming the Hessian into a positive-definite matrix as
\begin{eqnarray}
H(q)=\sum _{1\leq i\leq d} \lambda _i\psi _i\psi _i^T \mapsto G(q)=\sum _{1\leq i\leq d} g(\lambda _i)\psi _i\psi _i^T,
\end{eqnarray} 
where the differentiable function $g$ approximates the absolute value, i.e., $g(\lambda )\approx |\lambda |$. In the above, $\{ \lambda _i\} _{1\leq i\leq d}$ and $\{ \psi _i\} _{1\leq i\leq d}$ are the eigenvalues and eigenvectors of the Hessian $H$. Employing this transformed metric is equivalent to rescaling the motion of particles according to the absolute value of the curvature of the log-density's graph. Betancourt \cite{Betancourt2013a} further showed how to compute the right-hand sides of Eq.(\ref{d_Hamiltonian}) for the soft-absolute Hessian as follows: for $\Psi =\mathrm{mat}( \{ (\psi _i)_j\} )$,
\begin{eqnarray}
\Psi ^T\partial _{q_i}G(q)\Psi =T\odot (\Psi ^T\partial _{q_i}H(q)\Psi), \ \ T_{j\ell }=\left \{ \begin{array}{cc} \frac{g(\lambda _j)-g(\lambda _\ell )}{\lambda _j-\lambda _\ell } & (\lambda _j\neq \lambda _\ell ) \\ g^{\prime }(\lambda _j) & (\lambda _j=\lambda _\ell) \end{array}\right. , \ \ (1\leq i,j,\ell, \leq d), \label{d3_transform}
\end{eqnarray}
where $\odot $ denotes the Hadamard product. Given that eigenvectors and eigenvalues were already obtained, Betancourt \cite{Betancourt2013a} showed that the formula in Eq.(\ref{d3_transform}) allows one to compute the right-hand side of the second equivalence in Eq.(\ref{d_Hamiltonian}) for $1\leq i\leq d$ with an $O(d^3)$ computational cost by first caching
\begin{eqnarray}
W_1=\Psi BTB\Psi ^T, \ \ W_2=\Psi (R\odot T)\Psi ^T
\end{eqnarray}  
for $B=\mathrm{diag}((\Psi ^Tp)_i/g(\lambda _i))_i$ and $R=\mathrm{diag}(g(\lambda _i)^{-1})_i$, and then calculating 
\begin{eqnarray}
p^TG(q)^{-1}\frac{\partial G(q)}{\partial q_i}G(q)^{-1}p&=&tr\left (W_1\frac{\partial }{\partial q_i}H(q)\right ), \nonumber \\
tr\left (G(q)^{-1}\frac{\partial G(q)}{\partial q_i}\right )&=&tr\left (W_2\frac{\partial }{\partial q_i}H(q)\right ) . \label{Betancourt_formula}
\end{eqnarray}
Here, $W_1$ and $W_2$ are obtained by $O(d^{2.38})$ matrix multiplications, the full gradient $\frac{\partial }{\partial q}H(q)$ is obtained by an $O(d^3)$ computation, and the $2d$ trace operations are carried out by taking the sum of the elementwise products $2d$ times; hence, the overall complexity of the computation is $O(d^3)$.

\subsection{Structure-dependent removal of redundancy} \label{sec2.3}
As we show below, a straightforward application of the above algorithm to the model described by Eqs.(\ref{data_likelihood}), (\ref{BMKL}), (\ref{Solin_representation}) and (\ref{hpar_prior}) is time-consuming. We show that the computational cost can be reduced to $O(d^{\omega (r)}+d^{1+r})$ for $d=\mathrm{dim}(\{\{ a_{jkm}\} _{j,k,m}, \{ b_j\} _j\})+\mathrm{dim}(\{ \theta _{jk}\} _{j,k})$, $N=\lceil d^r \rceil$ and fixed $J$. Here, the exponent $\omega (r)$ denotes the computational complexity of multiplying matrices of sizes $d^r\times d$ and $d\times d$ (equivalently matrices of sizes $d\times d^r$ and $d^r\times d$) (see Table 2 and Footnote 1 of Ref.\cite{Gall2018}). This can be verified by substituting a concrete representation of the third derivatives of the negative log-posterior density for $\frac{\partial }{\partial q}H(q)$ in Eq.(\ref{Betancourt_formula}) as follows: for $s=1,2$, $a_{j}=\mathrm{vec} (\{ a_{jkm}\} _{(k,m)})$, $f_{j,i}=f_j(X_i)$, $\phi _{j, i}=\mathrm{vec} (\{ \phi _{jkm}(X_i)\} _{(k,m)})$, $\Phi _j =\mathrm{mat} (\{ \phi _{jkm}(X_i)\} _{i, (k,m)})$ and $W_{sj_1j_2}=\mathrm{mat} (\{ W_{s, j_1k_1m_1, j_2k_2m_2}\} _{(k_1,m_1),(k_2,m_2)})$,
\begin{eqnarray}
tr\left (W_s\frac{\partial }{\partial a_{j}}H(q)\right )&=&\sum _{i, j_1, j_2, k_1, k_2, m_1, m_2}\frac{\partial ^3U_i}{\partial _{f_{j_1, i}}\partial _{f_{j_2, i}}\partial _{f_{j, i}}}\phi _{j_1k_1m_1, i}\phi _{j_2k_2m_2, i}\phi _{j, i}W_{s, (j_1k_1m_1), (j_2k_2m_2)} +\cdots \nonumber \\
&=&\sum _{j_1, j_2} \Phi _{j}\left [ \mathrm{diag}\left ( \frac{\partial ^3U_i}{\partial _{f_{j_1, i}}\partial _{f_{j_2, i}}\partial _{f_{j, i}}}\right )  _i\left \{ \left ((\Phi _{j_1}W_{sj_1j_2})\odot \Phi _{j_2}\right )\mathrm{vec} (\{1\} )\right \} \right ]+\cdots,
\end{eqnarray} 
where we omit the computation that involves the derivatives with respect to the hyperparameters whose contribution could be computed in the same manner at a smaller computational cost. For each combination of $(j_1, j_2)$, the suitably ordered computation in the second line requires, $O(d^{\omega (r)})$, $O(d^{1+r})$, $O(d^{1+r})$, $O(d^r)$ and $O(d^{1+r})$ computations for calculating $\Phi _{j_1}W_{s,j_1,j_2}$, calculating the Hadamard product, calculating the matrix-vector product with vector $\mathrm{vec} (\{1\} )$ (a vector all of whose entries are $1$), multiplying the diagonal matrix from the left and calculating the matrix-vector product with $\Phi _{j}$, respectively.

The important point that should be noted here is that the standard library based on the automatic differentiation of the Hamiltonian does not follow the above order of computations. First, the direct derivatives of the Hamiltonian cannot be automatically computed when the soft-absolute metric is used. If the formula in Eq.(\ref{d_Hamiltonian}) is naively implemented by computing the third-order derivatives of the log-likelihood with the aid of an automatic differentiator in the reverse mode (the default mode in PyTorch), at least an $O(Nd^3)=O(d^{3+r})$ computation is required. This differs greatly from the efficient computation described above. Even if the joint likelihood is log-concave and the formula in Eq.(\ref{d_Hamiltonian}) can be avoided, a naive application of a reverse-mode automatic differentiator to the Hamiltonian results in a massive computational cost. In order to obtain an efficient implementation using an automatic differentiator, one must first implement a differentiator that recognises the formula in Eq.(\ref{d_Hamiltonian}) and then suitably assign the forward or reverse mode to the differentiation of each component of the Hamiltonian, according to the model structure.

\subsection{Dynamically programmed eigendecomposition} \label{sec2.4}
The formula given by Betancourt \cite{Betancourt2013a} relies on the availability of the complete set of eigenvectors $\Psi$. The fact that $H(q(t))$ and $H(q(t+\epsilon ))$ are close to each other motivates us to reduce computational costs by dynamically computing $\Psi (q(t+\epsilon ))$ to take advantage of $\Psi (q(t))$. The boundedness of the gradient of $H(q)$ and $p$ in the region where the joint probability for $(q,p)$ is concentrated implies that 
\begin{eqnarray}
\Psi (q(t))^TH(q(t+\epsilon ))\Psi (q(t))=\mathrm{diag}(\lambda _i(q(t)))+O(\epsilon )
\end{eqnarray}
holds with a high probability in terms of the Frobenius norm and can be efficiently eigendecomposed as
\begin{eqnarray}
\Psi (q(t))^TH(q(t+\epsilon ))\Psi (q(t))=Q(t+\epsilon, t)\mathrm{diag}(\lambda _i(q(t+\epsilon )))Q(t+\epsilon ,t)^T 
\end{eqnarray}
using the Jacobi method. Note that cyclic versions of the Jacobi method quadratically converge \cite{vanKempen1966} and are suitable for parallelisation \cite{Golub2013}, and thus are expected to carry out the above decomposition very quickly with a small error tolerance $\zeta $. With this decomposition, we update the eigenvectors as
\begin{eqnarray}
\Psi (q(t+\epsilon ))=\Psi (q(t))Q(t+\epsilon , t).
\end{eqnarray} 
In practice, we perform the Gram-Schmidt orthogonalisation of $\Psi (q(t))$ every ten steps before applying it to $H(q(t+\epsilon ))$ to remove accumulated numerical errors.
\subsection{Numerical experiments on computational complexity} \label{sec2.5}
We investigate the efficiency of the proposed algorithm by performing Bayesian logistic regression with artificially generated data and measuring its computation time. For comparison, we also perform posterior sampling using Hamiltorch \cite{Cobb2019}, a publicly available library for different types of HMCs based on PyTorch that works on CUDA devices. First, we generate standardised explanatory variables $X_i^{(\mathrm{exp})}\in \mathbf{R} ^D\overset{i.i.d.}{\sim }\mathcal{N} (0, 1)$ for each sample $i$ ($1\leq i\leq 500$). Then, we calculate the true log-odds ratio $f_1^*(X_i^{(\mathrm{exp})})$ according to 
\begin{eqnarray}
f_1^*(X_i^{(\mathrm{exp})})=\sum _{1\leq k\leq D}\sum _{m=1}^{M_{1k}} a_{1km}^*\phi _{1km}(X_{ik}^{(\mathrm{exp})})+b_1^*,
\end{eqnarray}
with $b_1^*=-0.5$, $a_{1km}^*\overset{i.i.d.}{\sim }\mathcal{N} (0, c^*m^{-1})$ for $4\leq m\leq 16$ and $a_{1km}^*=0$ otherwise. For any $k$, the feature function $\phi _{1mk}$ is given by Eq.(\ref{feature_functions}) in Appendix \ref{appendix_Solin} with $d=1$ and $L_1=8$. Here, the value of constant $c^*$ is determined so that the empirical standard deviation $\widehat{\mathrm{SD}} [f_1(X^{(\mathrm{exp})})]$ is 1.5. 

We then generate the sample label according to 
\begin{eqnarray}
X_i^{(\mathrm{lab})}=\left \{ \begin{array}{cc} +1 & \mathrm{prob.} \ \frac{1}{1+e^{-f_{1}^*(X_i^{(\mathrm{exp})})}}\\ -1 & \mathrm{otherwise} \end{array}\right. .
\end{eqnarray}
For this dataset, we perform posterior sampling, where the model is determined by Eq.(\ref{data_likelihood}) with $J=1$ and 
\begin{eqnarray}
U(f_1(\{ X_i^{(\mathrm{exp})}, X_i^{(\mathrm{lab})}\} ))=\ln (1+\exp (-X_i^{(\mathrm{lab})}f_1(X_i^{(\mathrm{exp})}))),
\end{eqnarray} 
where function $f_1$ is further decomposed into functions associated with one-dimensional Gaussian kernels: 
\begin{eqnarray}
f_1(X_i^{(\mathrm{exp})})=\sum _{1\leq k\leq D} \sum _{1\leq m\leq M_{1k}} a_{1km}\phi _{1km}(X_{ik}^{(\mathrm{exp})})+b_1.
\end{eqnarray}
Here, we abuse a notation for brevity by identifying index $k$ with a Gaussian kernel function $k(X_i, X_i^{\prime })=\exp (-(X_{ik}-X_{ik}^{\prime })^2/\sigma _{1k})$, for which, again, the feature function $\phi _{1mk}$ is given by Eq.(\ref{feature_functions}) in Appendix \ref{appendix_Solin} with $d=1$, $L_1=8$ and $M_{1k}=30$. The associated GPs and their hyperparameters are described by Eqs.(\ref{BMKL}), (\ref{Solin_representation}) and (\ref{hpar_prior}). For the transformation into the soft-absolute Hessian in the proposed method, we used $g(\lambda )=\sqrt{\kappa ^2+\lambda ^2}$, which takes values close to $g(\lambda )=\lambda \mathrm{coth} (\kappa ^{-1}\lambda )$ employed in Hamiltorch. We did not implement the latter because of the singularity of $\mathrm{coth}$ at $0$ and the lack of a straightforward implementation in the library we used (NVIDIA Inc., NVHPC23.1 \cite{NVHPC23.1}).  In addition to the values specified above, we used the following values: (for the comparison of sampling speed) $\alpha _\theta =\beta _\theta =2.0$ for any hyperparameter $\theta $, $\Sigma =1$, $A=9600$, $A_0=2400$, $C=100$, $\epsilon =0.001$, $\kappa =1$, $\zeta =1.0\times 10^{-13}$; (for the calculation of BME (see section \ref{appendix_BME})) $\alpha _{c_g}=\alpha _{c_\ell} =5, \ \beta _{c_g}=\beta _{c_\ell }=0.5,\ \alpha _{\sigma _g}=\beta _{\sigma _g}=1$, $\Sigma =1$, $A=50$, $C=100$, $\epsilon =0.001$, $\kappa =1$, $\zeta =1.0\times 10^{-13}$, $Z=50$, $\tau _s=1-0.02(s-1)$ ($1\leq s\leq 41$), $\tau _s=0.2-0.005(s-41)$ ($41<s\leq 71$), $\tau _s=0.05-0.002(s-71)$ ($71<s\leq 91$) and $\tau _s=0.01-0.001(s-91)$ ($91<s\leq 101$). The initial parameter values for MC we used were $a_{jkm} =0$, $b_j=0$ and $\theta =1$ for any hyperparameter $\theta $. We confirm the convergence of the proposed algorithm by performing Wilcoxon's rank-sum test for the values of the log-posterior density in the first and second halves of the trajectory after an initial burn-in period. The proposed method was implemented by writing codes in C++ and OpenACC that offload most of the calculation to a CUDA device retaining all necessary data in its own memory and invoke cuBLAS and cuSolver for linear-algebra routines as much as possible. For comparison, we wrote a code in PyTorch for the same model that invokes Hamiltorch for carrying out RMHMC or NUT-HMC with a manually coded likelihood. To conduct a comparison of different implementations, we measure the wall time spent on MC moves run on a single CPU core (a recent version of Intel Xeon processor) connected to a single NVIDIA Tesla A100 PCIe 80GB GPU card, for each of five datasets generated with different seeds for the random number generator.   
\subsection{Analysis of 1987 national medical expenditure survey (NMES)} \label{sec2.6}
To demonstrate that our implementation is acceptably efficient in a real-world application, we perform Bayesian estimation with excerpted data from the 1987 national medical expenditure survey (NMES) of the United States. A few groups of authors performed causal inference with this dataset about the effects of smoking on medical expenditures \cite{Johnson2003, Imai2004, Galagate2016}. In the present analysis, we attempt to increase the precision of the propensity function used for causal inference. As shown by Imai and van Dyk \cite{Imai2004}, the identification of a set of parameters $f$ that characterise the conditional distribution $p_{f}(X^{(\mathrm{tgt})}|X^{(\mathrm{cov})})$ of the actual treatment variable $X^{(\mathrm{tgt})}$ for the given values of covariates $X^{(\mathrm{cov})}$ reduces the dimensionality of subsequent causal analysis. In this dataset, $X^{(\mathrm{tgt})}$ denotes the packyear of smoking (i.e., the product of the number of packs of cigarette consumed by the subject and the duration of smoking measured in years). The covariates $X^{(\mathrm{cov})}$ include the age at the time of the survey, the age at the initiation of smoking, gender, race, marital status, education level, census region, poverty status and seat belt usage. The previous study focused mainly on the estimation of the mean of the conditional distribution as $f$ and not its variance. We compare the performance of linear and nonlinear models that describe only the conditional mean, and the performance of linear and nonlinear models that describe both the conditional mean and variance (Eq.(\ref{mvmodel})). In particular, as the model that best describes the given data is determined by the value of the Bayesian model evidence (BME) \cite{Bishop2006}, we investigate whether the proposed method computes this value within a reasonable computation time (see section \ref{appendix_BME} for the technical details of the calculation of BME). Carrying out causal inference with the estimated propensity functions requires further theoretical development and is not within the scope of the present study. Thus, we restrict our analysis to the estimation of propensity functions and discuss the further development that is needed in section \ref{sec4}.

We obtained the dataset included in a library for causal inference \cite{Galagate2016}. The categorical covariates in this dataset were transformed to a set of binary covariates that retained the original information. The continuous covariates were standardised to have a zero mean and unit variance.

Next, we consider the data likelihood described by Eq.(\ref{mvmodel}). The function $f_j$ ($j=1,2$) is described as 
\begin{eqnarray}
f_j(X_i^{(\mathrm{cov})})=\sum _{1\leq k\leq D} \sum _{1\leq m\leq M_{jk}} a_{jkm}\phi _{jkm}(X_{ik}^{(\mathrm{cov})})+b_j.
\end{eqnarray}
For the linear models, we use a linear kernel indexed by $k$ for each variable $X_{ik}^{(\mathrm{cov})}$, and thus we have $\phi _{jk1}(X_{i}^{(\mathrm{cov})})=X_{ik}^{(\mathrm{cov})}$ with $M_{jk}=1$. For the nonlinear models, we use a one-dimensional Gaussian kernel for each continuous variable with $L_1=8$ and $M_{jk}=30$, and a linear kernel for each binary variable. Models that describe the conditional mean and variance are determined in this manner. Models that describe only the conditional mean determine $f_1$ in the manner described above, whereas their variance is described by $f_2=b_2$. The priors for GPs and hyperparameters are described by Eqs.(\ref{BMKL}), (\ref{Solin_representation}) and (\ref{hpar_prior}) in the same manner as for the simulated data. In addition to the values specified above, we used the following values: $\alpha _{c_g}=\alpha _{c_\ell} =5, \ \beta _{c_g}=\beta _{c_\ell }=0.5,\ \alpha _{\sigma _g}=\beta _{\sigma _g}=1$, $\Sigma =1$, $A=50$, $C=400$, $\epsilon =0.0001$ ($\epsilon =0.00008$ for the nonlinear model describing both of the conditional mean and variance), $\kappa =1$, $\zeta =1.0\times 10^{-13}$, $Z=10$, $\tau _s=1-0.02(s-1)$ ($1\leq s\leq 41$), $\tau _s=0.2-0.005(s-41)$ ($41<s\leq 71$), $\tau _s=0.05-0.002(s-71)$ ($71<s\leq 91$) and $\tau _s=0.01-0.001(s-91)$ ($91<s\leq 101$). The initial parameter values for MC we used were $a_{jkm} =0$, $b_j=0$ and $\theta =1$ for any hyperparameter $\theta $.   
\section{Results} \label{sec3}
\subsection{Comparison among different implementations of RMHMC with a soft-absolute Hessian metric} \label{sec3.1}
The proposed implementation is fast enough to converge to equilibrium in both small and moderately large models, as seen from the trajectories in Fig.1(A) and confirmed by a Wilcoxon rank-sum test for $\sim 14,000$- and $\sim 60,000$-second simulations ($p=0.31, 0.47$), respectively. As shown in Fig.1(A) and (B), our implementation is roughly 10 times faster than the RMHMC based on Hamiltorch. To investigate the relative impact of the order of computations compared with that of the dynamically programmed eigendecomposition, we also measured the wall time for implementations in which the eigendecomposition was replaced by static ones based on either the Jacobi method or the divide-and-conquer algorithm. The inset in Fig.1(B) suggests that the difference in computation time between our implementation and the implementation based on Hamiltorch is mainly attributed to the computation of the gradient flow rather than eigendecomposition. However, a close examination of Fig.1(B) also shows differences among the algorithms for eigendecomposition. Although little difference is observed in computation time among the examined algorithms for small $D$, the dynamic eigendecomposition saves substantial computation time for large $D$. Examining the number of sweeps in the cyclic Jacobi method in static and dynamic implementations (Fig.1(C)), we confirm the advantage of the dynamic implementation, which requires only one to two sweeps at each step regardless of the model dimensionality, whereas the static implementation requires an increasing number of sweeps for greater model dimensionality on average.

\begin{figure}[H]
\includegraphics[width=120mm]{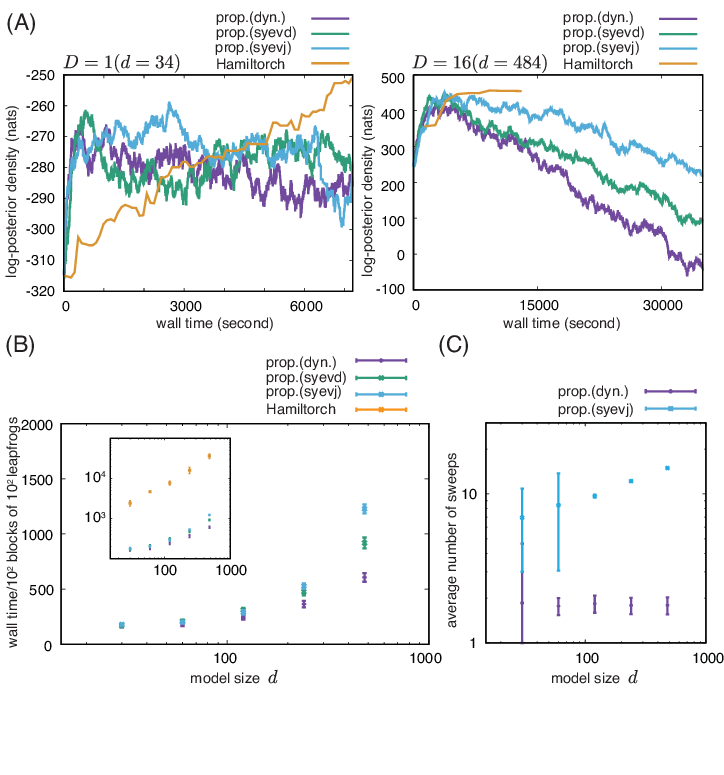}
\caption{RMHMC for Bayesian logistic regression on simulated data. Comparison between our implementations based on the dynamic and static eigendecomposition of the Hessian and the implementation based on Hamiltorch. (A) Representative trajectories of log-posterior density for $D=1$ ($d=34$) and $D=16$ ($d=484$). (B) Mean and standard deviation of the wall time spent on computing $100$ blocks of $100$ leapfrogs are shown on the linear (main panel) and logarithmic (inset) scales for different model sizes. (C) Mean and standard deviation of the number of sweeps in the cyclic Jacobi method for a single dynamic and static eigendecomposition of the metric are shown on the logarithmic scale for different model sizes. In (B) and (C), model sizes are shown on the logarithmic scales. (Abbreviations) prop.(dyn.): the proposed method based on dynamic eigendecomposition, prop.(syevd): the proposed method based on static eigendecomposition with the divide-and-conquer algorithm, prop.(syevj): the proposed method based on static eigendecomposition with the Jacobi method. }
\label{fig1}
\end{figure}

\subsection{Comparison with NUT-HMC sampler}  \label{sec3.2}
We also confirm the advantage of using RMHMC by comparing its performance with that of NUT-HMC. As shown in Fig.2A, NUT-HMC reaches samples of parameter values that have large values of log-posterior density very quickly and continues generating similar samples (Fig.2(A)). This contrasts with the slow convergence of RMHMC to generate samples with lower log-posterior density. We investigate which samplers generate representative samples from the posterior as follows. First, we examine the eigenvalues of the Hessian of the log-posterior density with representative samples from the two samplers (Fig.2(B)). We see that the eigenvalues for the sample from NUT-HMC are 10 times larger than those for the sample from RMHMC, except for several small positive or negative eigenvalues. This suggests that the samples from NUT-HMC have been taken from a much narrower region having a larger density. We therefore doubt that the total posterior probability for this narrow region is smaller than that for the region explored by RMHMC. Indeed, as we run RMHMC using for the initial parameter value a sample obtained using NUT-HMC, we observe that RMHMC swiftly moves to a region with the same lower range of log-density as that of the samples obtained by using the default initial condition $f_{jk}=0$ and $\theta _{jk\ell}=1$ (Fig.2(A)).

\begin{figure}[H]
\includegraphics[width=120mm]{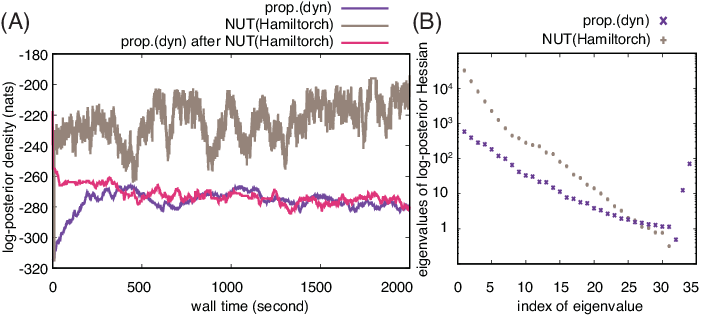}
\caption{Comparison of proposed implementation of RMHMC and NUT-HMC in Hamiltorch. (A) Representative trajectories of the log-posterior density for $D=1$ ($d=34$) obtained using the proposed implementation of RMHMC based on dynamic eigendecomposition (prop.(dyn)) and NUT-HMC in Hamiltorch from the same initial condition $f=0$ and $c_g, \sigma _g, c_\ell =1$. A representative trajectory for the proposed implementation of RMHMC starting with the parameter values obtained after 2000 seconds of sampling with NUT-HMC is also shown. (B) Eigenvalues of the Hessians of the negative log-posterior density for the samples of parameter values obtained after running the proposed implementation of RMHMC and NUT-HMC in Hamiltorch for 2,000 seconds. The eigenvalues are plotted in descending order on the logarithmic scale. For negative eigenvalues, their absolute values are plotted. The absolute values of the 32nd to 34th eigenvalues for the sample from NUT-HMC were extremely small and are not shown in the plot. }
\label{fig2}
\end{figure}

\subsection{Calculation of BME beyond the Laplace approximation with simulated data and NMES data}  \label{sec3.3}
The existence of negative eigenvalues in Fig.2(B) suggests that the posterior cannot be regarded as being approximately normal. This is not surprising, since hierarchical models are known to be often singular \cite{Watanabe2009}. Thus, the BME of the proposed model cannot be calculated by simply applying the Laplace approximation to the posterior density. Even in this case, we successfully carried out the calculation of BME through MC integration with reasonably high precision, as confirmed by the standard error of the estimate determined with multiple sequences of RMHMC [Table 1]. The obtained value was in good agreement with the BME value obtained by numerically integrating the Laplace approximation of the likelihood conditioned on each combination of hyperparameter values (see Appendix \ref{appendix_BME} for technical details). The latter calculation based on the Laplace approximation is justified by the fact that the model has log-concave posterior density and is regular when conditioned on hyperparameters.  

\begin{table}[!ht]
\centering
\begin{tabular}{lccccc}
\Xhline{1pt}
{\bf Data}& {\bf Model } & {\bf Estimated BME} &  {\bf Laplace Approx.} & {\bf Wall time (sec)} & {\bf \# MC ($Z$)} \\
\hline 
\ \ Simulation  & NL-logistic  & $-279.81\pm 0.40$ & $-282.31$ & $16243\pm 1102$ & 50 \\
\ \ NMES & L-mean & $-11716.78\pm 2.11$ & ---  & $43314\pm 5120$ & 10 \\
\ \ NMES & NL-mean & $-11602.37\pm 2.65$ & --- & $49648\pm 3613$ & 10 \\
\ \ NMES & L-mean/var & $-9540.63\pm 1.87$ & --- & $49424\pm 4509$ & 10 \\
\ \ NMES & NL-mean/var & $-9163.53\pm 3.43$ & --- & $52284\pm 3501$ & 10 \\
\Xhline{1pt}
\end{tabular}
\caption{Calculation of BME using the proposed implementation of RMHMC. The mean and standard error of the estimated model evidence and the mean and standard deviation of the wall time for its computation in a single RMHMC (among $Z$ RMHMCs) are shown. For simulated data, we also show the BME values estimated by iterating and integrating the Laplace approximation for different values of the hyperparameters. For NMES data, linear (L) and nonlinear (NL) models that describe only the conditional mean (mean) or both of the conditional mean and variance (mean/var) were used.} \label{table_BME}
\end{table}

To demonstrate the usefulness of the proposed method for analysing real-world data, we also calculate the BME for the NMES dataset using linear or nonlinear models that estimate the mean (and variance) of the distribution of the treatment variable conditioned on the covariates, namely, a propensity function. We successfully calculated BME values for this estimation problem within an acceptable computation time (Table 1). The computed BME values showed improved estimation of the propensity function with the nonlinear model for both the mean and variance of the conditional treatment density. 

\section{Discussion} \label{sec4}
In the present study, we have investigated MC methods for the posterior sampling of functions and hyperparameters in hierarchical GP models, and we have shown that a straightforward implementation of RMHMC based on currently available general-purpose libraries is highly redundant and its performance can be greatly improved. The main source of redundancy was the order of computation in the calculation of the gradient flow on the Riemannian manifold, whereas the eigendecomposition of the metrics was also a major source of redundancy for the larger models. These findings have non-trivial implications for future modelling based on GP and its implementations, because our results indicate that the current standard practice of coding only the likelihood of the model and allowing libraries to carry out inference with the aid of automatic differentiation results in poor performance, unless an intelligent library that optimises overall computational complexity is developed. This problem was not recognised in the previous study of RMHMC for GP \cite{Paquet2018}, because they sampled in the space for dual variables ($\{ f_{j,i}\} _{1\leq j\leq J,1\leq i\leq N}$ with $J=1$ in our notation) for which the entries of the third-order derivative tensor were sparse. In a hierarchical model with multiple GPs, the use of dual variables incurs a large computational cost. In this case, sampling in the space for multiple sets of dual variables $\{ f_{jki}\} _{1\leq j\leq J, k\in \mathcal{K} _j, 1\leq i\leq N}$ ($J>1$) is very high-dimensional. In this case, the eigendecomposition of the metric required for the Betancourt's formula (Eq.(\ref{d_Hamiltonian})) is essentially impossible to compute. Furthermore, the derivatives of posterior density with respect to both of the dual variables and the hyperparameters ($c_{jk}$ and $\sigma _{jk}$ in our notations) are also difficult to compute. The previous study avoided these difficulties by using only a single GP and fixing the hyperparameter values. In contrast we use a representation \cite{Solin2020} with a reduced number of primal variables, which makes the dependence on hyperparameters easier to compute.

In our numerical study, the dependence of computation time on model size does not precisely agree with theoretical expectations (section \ref{sec2.3}), presumably because the GPU accelerator carries out many arithmetic operations in parallel, and the effect of the size of multiplied matrices in the examined range is masked by this parallelism. Because this effect of parallelism is also observed in the multiplication of much larger matrices \cite{Dgemm2012}, we do not attempt to extend the range of dimension of our experiment. The computation time saved by dynamically programmed eigendecomposition was not striking in our simulation study; however, this effect could be much greater for larger model sizes, because eigendecomposition requires $O(d^3)$ computation, which eventually outweighs the computation of the gradient flow at sufficiently large $d$. We could not directly observe this phenomenon because tuning the step size and the threshold parameter for the soft-absolute Hessian metric becomes difficult for much larger $d$. We also note that the scaling of wall time for eigendecomposition depends heavily on the available GPU card. Several years ago, it was shown that the Jacobi method is outperformed by the divide-and-conquer algorithm for matrices larger than $512\times 512$ on a NVIDIA Tesla K40 GPU card \cite{Chien2017}; however, a recent study using dynamically programmed eigendecomposition for matrix optimisation showed that one sweep of the cyclic Jacobi method for matrices of size $4,096\times 4,096$ takes much less time than the divide-and-conquer algorithm for matrices of the same size \cite{Fawzi2021}. Because the dynamic programming successfully reduced the number of sweeps to fewer than two, regardless of model size, in our study, we expect that users can also benefit from the dynamically programmed eigendecomposition in larger models if a recent version of a GPU card is used.

In the numerical study, we showed that RMHMC outperforms NUT-HMC. The latter apparently reaches a parameter region with higher log-posterior density more quickly, but this region has turned out to be a narrow spurious region from which RMHMC swiftly moved away. Such entrapment is a well-known phenomenon and was the main motivation for the development of RMHMC \cite{Girolami2011}. However, various elaborations have been introduced to Euclidean HMC and related Langevin algorithms but not yet to RMHMC. For example, avoiding random-walk behavior with a no-U-turn mechanism was proposed for Euclidean HMC \cite{Hoffman2014} and RMHMC \cite{Betancourt2013b}, but it has not yet been properly implemented for the latter. The friction mechanism introduced to Euclidean Langevin MC \cite{Xiang2018, Arnak2020} suppresses inefficient oscillatory behaviour and could also be beneficial if applied to RMHMC. One factor that might have hindered elaboration of RMHMC in this regard is the previously poor performance of the plain RMHMC. Now that we have demonstrated improved RMHMC performance, further development in this direction should be encouraged.

Although we have successfully shown that RMHMC can be greatly accelerated, it is fair to note that we restricted our investigation to simple model settings. Bayesian multiple-kernel models have been shown to have favourable statistical properties when irrelevant GP is plugged out with sparsifying mechanisms such as a prior that imposes a penalty according to the number of included GPs \cite{Suzuki2012}. For this purpose, reversible-jump mechanisms \cite{Green1995, Karagiannis2013} must be introduced to RMHMC. Roughly speaking, this amounts to performing a shorter version of the MC integration in the calculation of BME multiple times in the simulation. For real-world applications, more elaborate structured models such as those for time-series data must sometimes be used. In this case, RMHMC for GP must be combined with other MC methods, such as sequential MC \cite{Moral2004, Chopin2020}. Whether the proposed method works within a reasonable computation time when it is tailored to the models described above needs to be investigated. Although the present work was intended to solve a biometrical and econometrical problem, unbiased effect estimation with Bayesian models requires further theoretical development, not just an accelerated implementation. The posterior we inferred with NMES data was asymptotically biased, as are essentially all machine-learning estimators. Therefore, in order to complete the causal inference with the biased propensity function, one needs to construct a corrected estimator for the treatment effect. This may be carried out by using the framework of doubly-robust debiased machine learning (DML) \cite{Chernozhukov2018} (see our previous work \cite{Hayakawa2025} for the application of DML to a model based on multiple reproducing kernel Hilbert spaces). However, its application is not straightforward, because the hierarchical Bayesian model is apparently singular (Fig.2(D)), whereas DML relies on the asymptotic normality of the estimators. The model is expected to be regular for fixed hyperparameter values and the framework of DML may be applied to a regular submodel conditioned on suitable hyperparameter values. As the focus of the present study is on the efficiency of RMHMC, we leave this development to a future work.

\section{Appendix}

\begin{table}[H] 
\caption{Mathmatical notations\label{table2}.  }
\begin{tabular}{|l|l|}
\hline
{\bf Symbols}& {\bf Description}  \\
\hline \hline 
$\mathbf{R} , \mathbf{R} ^d$ & The sets of real numbers and $d$-dimensional Euclidean space \\
$\lceil a \rceil$ & The smallest integer that is greater than or equals $a$ \\
$v^{T}$, $A^T$ & Transposition of vector $v$ and matrix $A$ \\
$(v)_i$ & The $i$-th element of vector $v$ \\
$(A)_{ij}$ & The element of matrix $A$ in the $i$-th row of the $j$-th column \\
$\mathrm{vec} (\{ v_i\} _i)$ \tablefootnote{Sometimes a combinatorial index such as $\mathrm{vec}(\{v_{km}\} _{(k,m)})$ is used. In this example, the index runs through all possible values for the combination $(k,m)$. } & Vector whose $i$-th element is $v_i$ \\
$\mathrm{mat} (\{ A_{ji}\} _{i,j})$ \tablefootnote{The first and second subscripts of the bracket $\{ \cdot \}$ specify the indices for the row and column, respectively.} & Matrix whose $i$-th row of the $j$-th column is $A_{ji}$ \\
$\mathrm{diag} (a_i)_i$ & Diagonal matrix whose $i$-th element is $a_i$ \\
$\mathrm{tr} A$ & the trace of matrix $A$ \\
$a\in \mathcal{A}$ & Element $a$ of a set $\mathcal{A}$ \\
$A \odot B$ & The Hadamard product of $A$ and $B$ \\
$|\mathcal{A}|$ & The number of elements in a set $\mathcal{A}$ \\
$E[\cdot ]$ ($E_{X\sim p}[\cdot ]$) & Expectation of the argument random variable (for the specified distribution)\\ 
$\mathrm{Var} [\cdot ]$ & Variance of the argument random variable \\
$\widehat{\mathrm{SD}} [\cdot ]$ & Empirical standard deviation of the argument variable \\
$\overset{\mathrm{def}}{=}$ & Equation defining the object on the left-hand side \\
$i.i.d.$, ($\overset{i.i.d.}{\sim }$) & Independently and identically distributed (objects drawn from the right-hand side)\\  
$\mathrm{Uniform} (a,b)$ & Uniform probability distribution over the open interval $(a,b)$ \\
$\mathrm{Normal} (\mu ,\Sigma)$ & Gaussian probability distribution with mean $\mu$ and (co)variance $\Sigma$ \\ 
$\mathrm{InvGamma} (\alpha ,\beta)$ & Inverse Gamma probability distribution with shape and scale parameters $\alpha, \beta $ \\
$\mathrm{Bernoulli} (p)$ & Bernoulli probability distribution (value $1$ with probability  $p$, or $0$ otherwise) \\
\hline
\end{tabular}
\end{table}

\subsection{Reduced-rank representation of GPs (Svensson et al. \cite{Svensson2016} and Solin and S\"arkk\"a \cite{Solin2020})} \label{appendix_Solin}
For a $d$-dimensional translation-invariant isotropic kernel function $k(x, x^{\prime })$ describing the covariance of a GP, Solin and S\"arkka\"a introduced the following approximation in a rectangular domain $[-L_1, L_1]\times \cdots \times [-L_d, L_d]$: 
\begin{eqnarray}
k(x, x^{\prime })\approx \sum _{m_1,\cdots ,m_d=1}^{\hat{M} ^d} S(\sqrt{\lambda _{m_1\cdots m_d}})\phi _{m_1\cdots m_d}(x)\phi _{m_1\cdots m_d}(x^{\prime}) \label{Solin_approximation}
\end{eqnarray}
with the eigenfunctions and eigenvalues
\begin{eqnarray}
\phi _{m_1\cdots m_d}(x)=\prod _{\ell =1}^d\frac{1}{L_{\ell }}\sin \left (\frac{\pi m_\ell (x_\ell +L_\ell )}{2L_\ell }\right ), \label{feature_functions}
\end{eqnarray}
and 
\begin{eqnarray}
\lambda _{m_1,\cdots , m_d}=\sum _{\ell =1}^{d} \left (\frac{\pi m_\ell }{2L_\ell }\right ) ^2.
\end{eqnarray}
In the above, $S(\cdot )$ is the spectral density of the GP related to the kernel function $k(r)=k(x, x+r)$ via the Wiener-Khinchin theorem: 
\begin{eqnarray}
k(r)&=&\frac{1}{(2\pi )^d} \int _{\mathbf{R} ^d}S(\omega )e^{\mathrm{i} \omega ^Tr}d\omega, \nonumber \\
S(\omega )&=& \int _{\mathbf{R} ^d}k(r)e^{-i\omega ^Tr}dr.
\end{eqnarray}

In the present study, we focus on the one-dimensional ($d=1$) case. Then, the GP is represented in the form of Eq.(\ref{Solin_representation}). We also refer readers to Svensson et al. \cite{Svensson2016} for more details about its implementation. Unlike Svensson et al. \cite{Svensson2016}, we used scalar-valued GPs, and thus, the matrix normal distributions and inverse-Wishart distributions they used simply reduces to normal distributions and inverse-Gamma distributions in our case. By applying the Fourier transform to $k(r)=\exp (-r^2/\sigma )$, we obtain $S(\omega )=\sqrt{\pi \sigma }\exp (-\sigma \omega ^2/4)$. Considering also Eq.(\ref{Solin_approximation}), we have 
\begin{eqnarray}
V_{jkm}=\sqrt{\pi \sigma } \exp \left (-\frac{\pi m^2\sigma }{16L^2}\right ),
\end{eqnarray}
and 
\begin{eqnarray}
\phi _{jkm} (X_i)=\frac{1}{L}\sin \left (\frac{\pi m(X_{ik} +L)}{2L}\right ).
\end{eqnarray}
In practice, we dropped $\frac{1}{L}$ from $\phi _{jkm}$, which amounts to rescaling of $c_g$.

\subsection{MC integration for the calculation of BME (Calderhead and Girolami \cite{Calderhead2009})} \label{appendix_BME}
BME for the model with parameters $a=\{\{ a_{jkm}\} _{j,k,m}, \{ b_j\} _j\} $ for approximately describing $\{ f_{j}\} _{j}$ and hyperparameter $\theta =\{ c_g, \sigma _g, c_\ell \} $ (with prior densities $\mathcal{G}_\theta (a)$ and $\Pi (\theta )$) is defined as the following marginal likelihood \cite{Bishop2006}: 
\begin{eqnarray}
\mathrm{BME} \overset{\mathrm{def}}{=}\int P(X|a)\mathcal{G} _\theta (a)\Pi (\theta )dad\theta . \label{BME}
\end{eqnarray}
In a hierarchical model, carrying out the above integration is usually intractable. If the joint density is approximately Gaussian, the following Laplace approximation can be used: for $v=\mathrm{vec}(a, \theta)$,
\begin{eqnarray}
\mathrm{BME} &\approx & \int \widehat{P}\exp (-(v-\widehat{v})^T\widehat{H} (v-\widehat{v} ))dv \nonumber \\
&=&\widehat{P} (2\pi )^{d/2}|\widehat{H}|^{-1/2}
\end{eqnarray} 
where $\widehat{v}$ denotes the maximum-a-posteriori value of $v$, and $\widehat{P}$ and $\widehat{H}$ denotes the peak value and the negative Hessian of the logarithm of the joint density at $\widehat{v}$.

The above formula cannot be used when the deviation of the joint density from the Gaussian approximation is large. This is often the case when we use a singular (or nearly singular) hierarchical model \cite{Watanabe2009}. If the model is parametric, asymptotic formula for the BME of singular models can be used \cite{Watanabe2013, Drton2017}. However, these formula do not apply to the semiparametric models that we consider in this study.

Even in this case, the BME can be calculated by performing the following integration \cite{Calderhead2009}: 
\begin{eqnarray}
\mathrm{BME} =\int _{0}^{1} \mathrm{E} _{v\sim P_\tau }\left [\ln P(X|a)\right ]d\tau , \label{MC_integration}
\end{eqnarray}
with a set of probability densities parameterised by $\tau $ ($0\leq \tau \leq 1$):
\begin{eqnarray}
P_\tau (v)\propto P(X|a)^\tau \mathcal{G} _\theta (a)\Pi (\theta ).
\end{eqnarray}
This integration can be carried out by sampling from $P_\tau $ for each value of $\tau $ in discretised steps interpolating $0$ and $1$ and approximating the integrand with the statistics over the samples.

We carry out the above integration with multiple RMHMC indexed by $z=1,2,\ldots ,Z$ each of which determines the parameter value $(a_s^{(z)}, \theta _s^{(z)})$ used for the calculation of the integrand $\mathrm{E} _{v\sim P_\tau }[\ln P(X|a)]\approx \frac{1}{Z} \sum _z\ln P(X|a_s^{(z)})$ in Eq.(\ref{MC_integration}) for $\tau =\tau _s$  ($s=1,2,\ldots , S$; $\tau _1=1$ and $\tau _S=0$) after performing $A$ blocks of $C$ leapfrogs from the initial parameter value $(a_{s-1}, \theta _{s-1})$. To determine $(a_1, \theta _1)$ for $\tau _1=1$, we performed a single RMHMC until convergence and obtained a shared initial condition for $Z$ RMHMCs that perform further $500$ blocks of $400$ leapfrogs to obtain $(a_1^{(z)}, \theta _1^{(z)})$ from this initial condition.

For the above calculation, it should noted that, if $\tau _s-\tau _{s-1}$ is small enough for all $s$ and the number of MC moves $A$ for each $\tau _s$ is large enough, 
\begin{eqnarray}
\mathrm{BME} \approx \sum _{s=2}^{S} \frac{1}{2}\left (\ln P(X|a_s^{(z)})+\ln P(X|a_{s-1}^{(z)})\right )(\tau _s-\tau _{s-1})
\end{eqnarray} 
holds for each value of $z$. In this case, since $A$ is large enough, $\{ a_s\} _{1\leq s\leq S}$, can be considered independent. Assuming this independence, we have
\begin{eqnarray}
\mathrm{Var} [\mathrm{BME}] &\approx &\sum _{s=2}^{S-1} \frac{1}{4} \mathrm{Var} [\ln P(X|a_s^{(z)})]\{ (\tau _s-\tau _{s-1})^2+(\tau _{s+1}-\tau _s)^2\} \nonumber \\
&&+  \frac{1}{4} \left \{  \mathrm{Var} [\ln P(X|a_1^{(z)})] (\tau _2-\tau _1)^2+\mathrm{Var} [\ln P(X|a_S^{(z)})] (\tau _S-\tau _{S-1})^2\right \} ,
\end{eqnarray}
and its right-hand side vanishes as the discretisation of the integration interval becomes infinitely finer.

The use of multiple RMHMCs (that is, $Z>1$) is expected to accelerate the convergence. In practice, we used $A=50$ and $Z=10$ and confirmed that the change in the values of $A$ and $Z$ ($A=12, 25, 50, 100$ and $Z=5, 10, 20$) does not affect the decision about the best model.

Since $P(X|a)\mathcal{G} _{\theta }(a)$ is log-concave for fixed $\theta $, in the numerical experiment with simulated data, we also perform the following integration:
\begin{eqnarray}
\mathrm{BME} &\approx & \int \widehat{P} _\theta \exp (-(a-\widehat{a}(\theta ))^T\widehat{H} _{aa}(\theta )(a-\widehat{a} (\theta )))da\Pi (\theta )d\theta \nonumber \\
&=&\int \widehat{P} _\theta (2\pi )^{d/2}|\widehat{H} _{aa}(\theta )|^{-1/2} \Pi (\theta )d\theta ,
\end{eqnarray} 
where $\widehat{a}(\theta )$ denotes the value of $a$ that maximises $P(X|a)\mathcal{G} _\theta (a)$ for the given value of $\theta $, and $\widehat{P} _\theta $ and $\widehat{H} _{aa}(\theta )$ denote the peak value and the negative Hessian of $\ln P(X|a)\mathcal{G} _\theta (a)$ at $\widehat{a} (\theta )$. The two-dimensional integration in the second line was performed by discretising the rectangular region $[0, 4]\times [0, 4]$ with a mesh size $0.01\times 0.02$. For the discrete values of hyperparameters, we obtained $\widehat{a} (\theta )$ by performing limited-memory BFGS \cite{Liu1989} using PyTorch. Since we did not use linear kernels for simulated data, we can simply ignore the hyperparameter for linear kernels.

\section*{Acknowledgements} 
We would like to express our gratitude to two medical IT companies, 4DIN Ltd. (Tokyo, Japan) and Phenogen Medical Corporation (Tokyo, Japan) for financial support. Neither company had a role in the research design, analysis, data collection, interpretation of data, or review of the manuscript, and no honoraria or payments were made for authorship. 
\section*{Author contributions}
T.H. conceptualised the study. T.H. and S.A. designed models and simulation data. T.H. developed all program codes. T.H. and S.A. interpreted the analysed results. T.H. wrote the manuscript. T.H. and S.A. have read and agreed to its contents and have approved the final manuscript for submissions.
\section*{Code availability}
All of the source codes that support the findings of this study will be made available as supplementary materials upon acceptance to a journal.
\section*{Funding}
This research was supported by the Ministry of Education, Culture, Sports, Sciences and Technology (MEXT) of the Japanese government and the Japan Agency for Medical Research and Development (AMED) under grant numbers JP18km0605001 and JP223fa627011.
\section*{Conflict of interest}
The authors declare no conflicts of interest.

\bibliography{ref}
\end{document}